\documentclass[twoside,11pt]{article}

\usepackage{blindtext}

\usepackage[abbrvbib, preprint]{jmlr2e}

\usepackage[preprint]{jmlr2e}
\usepackage{lipsum,booktabs}
\usepackage{amsmath,mathrsfs,amssymb,amsfonts,bm,enumitem}
\usepackage{xcolor}
\usepackage[T1]{fontenc}
\allowdisplaybreaks
\usepackage{appendix}
\usepackage{mathtools}
\usepackage{multirow}
\usepackage{algorithmic,algorithm}

\newcommand\blfootnote[1]{%
  \begingroup
  \renewcommand\thefootnote{}\footnote{#1}%
  \addtocounter{footnote}{-1}%
  \endgroup
}

\usepackage{lastpage}

\firstpageno{1}

\begin{document}

\title{On Realization of Intelligent Decision-Making in the Real World: A Foundation Decision Model Perspective}

\author{\name Ying Wen$^{1}$ \email ying.wen@sjtu.edu.cn \\
\name Ziyu Wan$^{1}$\email alex\_wan@sjtu.edu.cn\\
\name Ming Zhou$^{1}$ \email mingak@sjtu.edu.cn\\
\name Shufang Hou$^{2}$ \email shufang.hou@digitalbrain.cn\\
\name Zhe Cao$^{1}$ \email caozheoldpaws@sjtu.edu.cn \\
\name Chenyang Le$^{1}$ \email nethermanpro@sjtu.edu.cn \\
\name Jingxiao Chen$^{1}$ \email timemachine@sjtu.edu.cn \\
\name Zheng Tian$^{3}$ \email zheng.tian@shanghaitech.edu.cn \\
\name Weinan Zhang$^{1}$ \email wnzhang@sjtu.edu.cn \\
\name  Jun Wang$^{2,4}$ \email jun.wang@cs.ucl.ac.uk  \\
\addr $^{1}$Shanghai Jiao Tong University \\
\addr $^{2}$Digital Brain Laboratory \\
\addr $^{3}$ShanghaiTech University \\ 
\addr $^{4}$University College London \\
}

\editor{}

\maketitle

\begin{abstract}
The pervasive uncertainty and dynamic nature of real-world environments present significant challenges for the widespread implementation of machine-driven Intelligent Decision-Making (IDM) systems. 
Consequently, IDM should possess the ability to continuously acquire new skills and effectively generalize across a broad range of applications. 
The advancement of Artificial General Intelligence (AGI) that transcends task and application boundaries is critical for enhancing IDM. 
Recent studies have extensively investigated the Transformer neural architecture as a foundational model for various tasks, including computer vision, natural language processing, and reinforcement learning. 
We propose that a Foundation Decision Model (FDM) can be developed by formulating diverse decision-making tasks as sequence decoding tasks using the Transformer architecture, offering a promising solution for expanding IDM applications in complex real-world situations. 
In this paper, we discuss the efficiency and generalization improvements offered by a foundation decision model for IDM and explore its potential applications in multi-agent game AI, production scheduling, and robotics tasks. 
Lastly, we present a case study demonstrating our FDM implementation, DigitalBrain (DB1) with 1.3 billion parameters, achieving human-level performance in 870 tasks, such as text generation, image captioning, video game playing, robotic control, and traveling salesman problems. 
As a foundation decision model, DB1 represents an initial step toward more autonomous and efficient real-world IDM applications.
\blfootnote{Work completed when Z. Wan, M. Zhou, Z. Cao, C. Le, J. Chen are interns at Digital Brain Laboratory. Correspondence to: Weinan Zhang <wnzhang@sjtu.edu.cn>.}
\end{abstract}

\begin{keywords}
  artificial intelligence, intelligent decision-making, transformer, foundation decision model
\end{keywords}

\section{Introduction}
Intelligent Decision-Making (IDM) leverages artificial intelligence (AI) and related technologies to address real-world decision-making tasks. Human production activities, from operating machinery to governing nations, rely on a series of decisions to achieve desired objectives. Decisions are made based on knowledge, experience, and intuition~\citep{khatri2000role}. 
However, the human mind's capacity is finite, limiting decision bandwidth for numerous complex and counter-intuitive problems. 
The emergence of digital technologies and information processing techniques has substantially increased the complexity and uncertainty of decision problems~\citep{peres2020industrial}, which often exceed human decision-making capabilities.
AI facilitates data analysis beyond human capacity, addressing the issue of 'bounded rationality'\citep{simon1990bounded}. 
Traditional IDM models, based on rules, data analysis, control theory\citep{kirk2004optimal}, and operations research (OR)\citep{winston2004operations}, enhance decision efficiency by automating aspects of the decision process in various domains\citep{eom2006survey}, such as production scheduling~\citep{herrmann2006handbook}, air traffic control~\citep{nolan2011fundamentals}, disease diagnosis~\citep{barnett1987dxplain}, and enterprise management~\citep{ranjan2009business}. 
However, due to low model complexity and the need for manual feature design and model selection, traditional IDM models are limited to solving specific tasks.

Existing IDM models primarily address narrowly defined decision-making tasks~\citep{eom2006survey} to support human decision-making. Given a fixed, small-scale, white-box scheduling problem, OR algorithms can generate a logistics scheduling strategy~\citep{gudehus2012comprehensive}. However, OR algorithms must re-perform the optimization for new scheduling problems without reusing historical experience. Consequently, it is challenging for existing decision-making models to provide an optimal strategy for more extensive, dynamic, and complex tasks with increased uncertainty and exponentially growing solution spaces, such as urban traffic light control~\citep{jensen2016vision} and state grid dispatch. These complex decision problems are common in real-world applications, which highlights the challenges in applying IDM to enhance operational capabilities and value in production activities.
  \begin{figure}[t!]
    \centering
    \includegraphics[width=1.\linewidth]{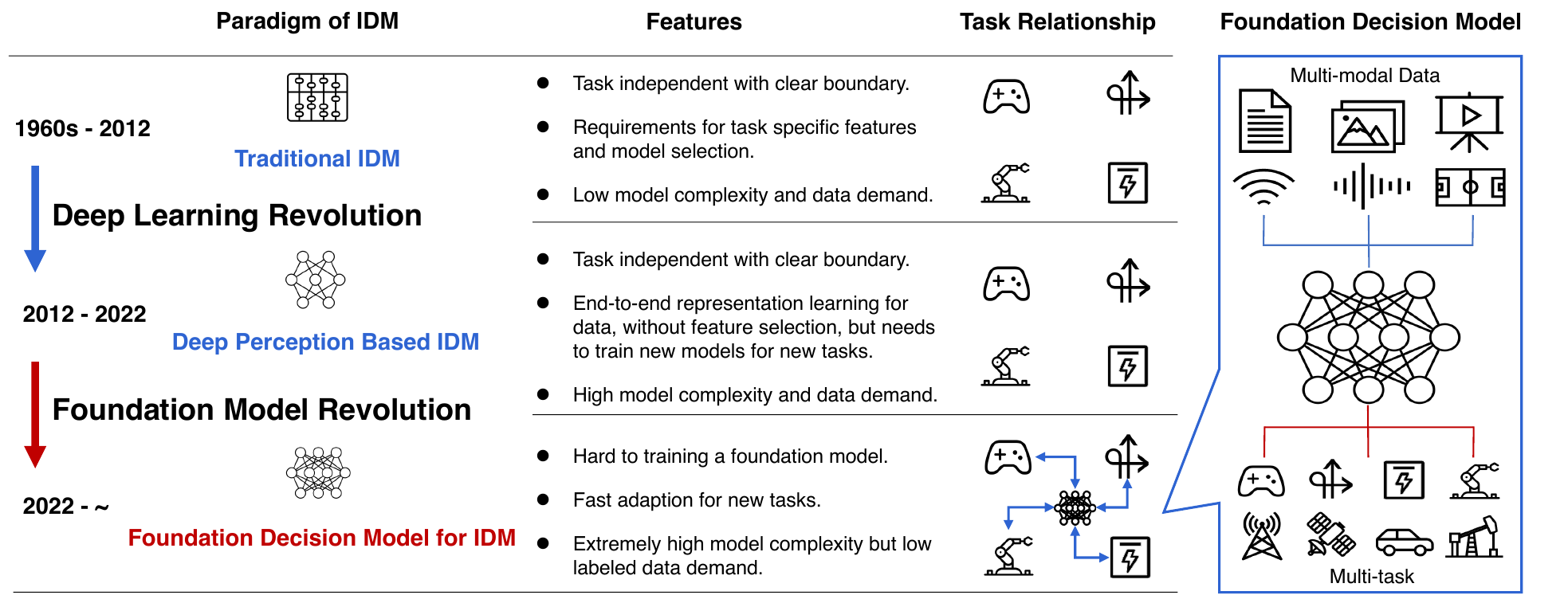}
    \caption{Paradigm shifts of IDM: from traditional data analysis, control theory and OR methods,  deep learning based perception methods to foundation model based methods.}
    \label{fig:fdm}
  \end{figure}

In the AI boom era, humans expect IDM to further improve efficiency, reduce costs, and find broader applications in complex real-world production activities~\citep{phillips2008intelligent}. AI has already demonstrated its potential to solve prediction tasks to accelerate automated decisions in business~\citep{gupta2020deep}. For example, an AI algorithm can predict whether a user will like a product and provide an estimated bidding price for a potential advertising slot~\citep{wang2017display}. Based on deep learning~\citep{lecun2015deep}, such IDM is more concerned with perception problems and excels at recognizing data patterns. Unlike traditional IDM models, deep perception-based IDM models employ end-to-end representation learning and high model complexity but require substantial amounts of data. Understanding data patterns can help reduce uncertainties during the decision process. However, it is not the ultimate goal of IDM. A complete decision cycle includes perceiving the environment and adjusting behaviors based on feedback.
Aligned with IDM's long-term vision to make successful decisions in the real world, AI aims to study intelligent agents that receive perceptions from the environment and learn to make adaptive decisions based on feedback to maximize expected utility~\citep{russell2010artificial}. Learning and feedback mechanisms are AI's key features, enabling the implementation of closed-loop IDM models in the next stage, which continuously adapt to more dynamic and complex environments, such as robotics~\citep{murphy2019introduction} and autonomous unmanned systems~\citep{chen2022unmanned} in real world settings.

The focus of IDM is now shifting from intelligent perception to higher-level autonomous decisions capable of interacting with and learning from the real world. Recent advances in deep reinforcement learning~\citep{mnih2013playing,sutton2018reinforcement} demonstrate that IDM can surpass human-level performance in multi-step complex decision-making tasks with well-defined problem settings, such as Go~\citep{silver2017mastering} and StarCraft 2~\citep{vinyals2019grandmaster}. However, IDM models are still in the early stages of widespread real-world application~\citep{sutton2019bitter}, as they can only solve specific, well-defined problems.

When applying IDM models to interact with real world environments, several unique challenges remain unaddressed, particularly efficiency and generalization issues. The real world consists of various tasks, goals, and agents in a constantly changing, dynamic environment, continuously generating new tasks and objectives~\citep{peres2020industrial}. Consequently, IDM applications face numerous practical issues, such as limited data, imbalanced class distributions, multi-modal data indexed over space and time, the need to carefully compose different models in multiple decision stages, and high-cost trial and error for decisions~\citep{gupta2020deep,qin2021neorl}. To tackle these challenges and improve existing IDM models, two questions must be answered: (1) Can data inputs/outputs be standardized with a unified interface? (2) How can the infinite number of real-world problems be mapped to tractable computational problems?

In recent years, transformer-based foundation models have gained popularity as unified solutions for various downstream tasks, including natural language processing (NLP)\citep{nlp} and computer vision (CV)\citep{computer-vision}. Notable examples include BERT~\citep{bert}, GPT-3~\citep{brown2020gpt3}, and the Swin Transformer~\citep{SwinTransformer}. Due to their high model complexity and low demand for labeled data, foundation models exhibit strong generalization capabilities for new tasks, becoming the new backbone network for text, image, voice, video, and related cross-modality tasks. Although foundation models were originally designed for prediction and generative tasks, it is worth exploring whether they can serve as effective neural architectures for decision-making problems. Recently, foundation models have demonstrated noticeable effectiveness and generalization ability in reinforcement learning (RL)\citep{sutton2018reinforcement}, such as the Decision Transformer (DT)\citep{decision-transformer}, Trajectory Transformer (TT)\citep{trajectory-transformer}, and Video Pre-training (VPT)\citep{baker2022vpt}, marking important strides towards general decision-making models.

Amidst the ongoing foundation model revolution, as illustrated in Figure~\ref{fig:fdm}, one might wonder if intelligent decision-making (IDM) models can culminate in a single transformer, the Foundation Decision Model (FDM), which treats various decision problems as sequence decoding tasks using transformer architecture to govern all decision-making tasks. FDM employs a standard transformer-based autoregressive decoder for decoding decisions, supporting multi-modality, multi-task, large-scale, and general-purpose decision-making tasks. FDM differs from the DT~\citep{decision-transformer} and TT~\citep{trajectory-transformer}, which utilize well-designed transformer architectures for reinforcement learning problems but lack extensibility for other decision-making tasks. The most closely related work is DeepMind's Gato~\citep{reed2022generalist}, a 1.2 billion parameter FDM that serializes all data into a sequence of tokens. Gato demonstrates that a single transformer can directly handle hundreds of tasks in robotics, single-agent games, and vision and language. As the first attempt at FDM, Gato has yet to reveal all possibilities of FDM. In this paper, we discuss the opportunity to use FDM as an IDM solution in real-world applications, significantly alleviating the poor sample efficiency and generalization issues of IDM models.

FDM generalizes to more decision problems by enhancing its ability to encode implicit knowledge and process multi-task~\citep{han2020survey}, multi-modal~\citep{xu2022multimodal,tsai2019multimodal}, and space-time data~\citep{wen2022transformers}. With the help of transformers, FDM integrates data across modalities, including time series, text, voice, image, video, categorical, or numerical data~\citep{bommasani2021opportunities,bao2021beit,BEIT3}, enabling a wide range of multi-agent applications, such as video games, robotics, and vehicle routing tasks. Moreover, FDM offers better training and inference efficiency than recurrent neural network-based models for modeling series and sequence patterns, due to their parallel computation in training or inference~\citep{bert}. Additionally, the self-attention mechanism in FDM can learn interactions between input tokens, inherently permutation invariant~\citep{lin2017structured} for processing a sequence of states and actions, making it well-suited for multi-agent tasks~\citep{wen2022multi}. Overall, FDM represents inputs as token sequences followed by self-attention interactions, intrinsically compatible with more modalities and tasks with strong generalization potential.

FDM addresses online decision problems in feasible time and cost domains more efficiently. Firstly, thanks to large-scale parameters, multi-head operators, and global aggregation~\citep{bert, brown2020gpt3}, FDM naturally possesses extremely high model complexity and learns cross-domain knowledge through effective pre-training on large-scale, multi-modal, and multi-task datasets. Secondly, FDM intrinsically shares the same learning objective across offline and online training stages, reducing the difficulty of continuous learning. Finally, FDM adopts tuning modes from the transformer, such as fine-tuning and prompt learning, further enhancing learning efficiency. In summary, FDM utilizes collected offline data for pre-training, significantly improving data efficiency and accelerating the calculation speed of online decision-making tasks.

We believe FDM has great potential as a unified solution by converting all types of tasks into sequence modeling problems, which is one of the promising directions toward realizing IDM in the real world by trading off efficiency and generalization.
We also discuss some possibilities of FDM in real world applications, including game AI, production scheduling, and robotics.
To verify our hypothesis, we develop DigitalBrain (DB1) with around 1.3 billion parameters.Â 
DB1 achieves human-level performance over 453 tasks, including text, text-image, video games, robotics, and path planning tasks.
DB1 is a new attempt at FDM and has extensions in parameter size, sequence length, and supported tasks compared to Gato.
However, DB1 is still a partial implementation of general FDM.
We indicate the gaps between DB1 and general FDM in Table~\ref{tab:db_comp}.
We believe DB1 is a further step towards general FDM, which points out the further improvement directions of FDM.

We believe FDM has tremendous potential as a unified solution by converting all types of tasks into sequence modeling problems, representing one of the most promising directions towards realizing IDM in the real world while balancing efficiency and generalization. We also discuss some possibilities of FDM in real-world applications, including game AI, production scheduling, and robotics. To verify our hypothesis, we develop DigitalBrain (DB1) with around 1.3 billion parameters. DB1 achieves human-level performance across 870 tasks, including text, text-image, video games, robotics, and path planning tasks. DB1 represents a new attempt at FDM with extensions in parameter size, sequence length, and supported tasks compared to Gato. However, DB1 is still a partial implementation of general FDM. We indicate the gaps between DB1 and general FDM in Table~\ref{tab:db_comp}. We believe DB1 is a further step towards general FDM, pointing out directions for future improvement of FDM.



\begin{table}[t!]
\caption{Comparisons between DB1 and general FDM models. As a partial implementation of general FDM, DB1 has 1.32B parameters and 2048 input sequence length, which supports text, image, category, numerical inputs and dialogue, image captioning, video games, robotics, and path planning tasks. DB1 uses offline training and supports fine-tuning for new tasks.}
\centering
\begin{tabular}{lll}
\toprule
                        & DB1      & General FDM                       \\
\midrule
Task support        & 5   & 5+                  \\
Data Modality          & 4       & 4+                          \\
Model parameter        & 1.2B      & 10B+                        \\
Sequence Length       & 2048        & 2048+                          \\
Training paradigm     & Offline       & Offline + Online         \\
Task Planning          & N/A        & Yes                       \\
Tuning model     & Fine-tuning      & Fine-tuning + Prompt      \\
\bottomrule
\end{tabular}%
\label{tab:db_comp}%
\end{table}%

In the rest of this paper, Section~\ref{sec:gen} focuses on the generalization possibility of multi-modal, multi-task decision-making tasks via a unified model.
Section~\ref{sec:efficient} discusses the efficiency issue of IDM algorithms to solve different types of decision problems,
Section~\ref{sec:app} introduces some typical applications,
Section~\ref{sec:case} demonstrates the potential of the FDM via a case study of the DB1 model, and
Section~\ref{sec:con} concludes the paper with a summary.

\section{FDM Makes Generalizable Decisions}
\label{sec:gen}
Real-world situations are rarely identical, making generalization essential for decision-making. FDM learns generalizable representations for multi-modal and multi-task decision-making problems, a fundamental requirement for IDM models to continuously adapt to new tasks~\citep{seger2013categorization,kirk2021survey}. Additionally, FDM can perform online adaptation after offline pre-training, further enhancing its generalization ability for life-long decision learning~\citep{poquet2021developing}.

\subsection{Multi-modal Decisions}
Originally designed for NLP~\citep{bert}, transformers have been extended to other modality tasks~\citep{xu2022multimodal}, allowing FDM to operate across diverse domains and modalities. Real world environments are multi-modal, meaning both observations and behaviors for IDM applications are multi-modal. For instance, a robot requires multi-modal sensors, such as cameras, voice, radar, and maps, to perceive the environment. Various deep learning methods have been proposed to enhance AI models' multi-modal perception ability. MultiModel~\citep{kaiser2017one} is trained jointly on eight distinct speech, image, and text processing tasks, using modality-specific encoders for text, images, audio, and categorical data, while the rest of the network parameters are shared across tasks. In 2018, the Google AI team proposed "one big net for everything"~\citep{schmidhuber2018one}, describing a method for incrementally training an increasingly general problem solver.

With the development of pre-trained foundation models, FDM is further equipped to interact with multi-modal environments. Research on autonomous IDM in multi-modal virtual environments~\citep{gan2020look} has emerged, particularly in visual navigation and object manipulation. In April 2022, the Google Robotics team released SayCan~\citep{ahn2022can}. SayCan robots generate sub-task instructions for a given task using a natural language pre-trained model and perform corresponding actions based on the sub-task to complete the goal. Thus, SayCan robots can solve tasks given natural language descriptions through visual perception in real environments. Gato~\citep{reed2022generalist}, launched by DeepMind in June 2022, unifies multi-step decision-making tasks, multi-round dialogues, and image-text generation tasks into a transformer-based autoregressive problem, achieving promising results in over 600 tasks. This preliminarily verifies the potential of multi-modal pre-trained foundation models for agent learning tasks. In July 2022, based on the open-world game Minecraft~\citep{perez2019multi}, NVIDIA's MineDojo~\citep{fan2022minedojo} provides multi-modal data, such as game videos, Wikipedia, and forum pages, and creates thousands of single-agent tasks. MineDojo leverages the CLIP pre-training model~\citep{radford2021learning} to conduct comparative learning on video-text pair datasets and train a zero-shot prediction model of the correlation between natural language targets and environmental visual states.

\subsection{Multi-task Decisions}
FDM is an end-to-end multi-task learning transformer framework that simultaneously learns multiple language, vision, and decision-making tasks. Multi-task learning (MTL)~\citep{vandenhende2021multi} seeks to utilize useful information from multiple tasks to improve the generalization performance of all tasks. Existing MTL studies primarily focus on supervised learning tasks, with few addressing tasks such as self-supervised and reinforcement learning tasks~\citep{zhang2021survey}. FDM jointly learns multiple tasks, including reinforcement learning and OR tasks. Moreover, FDM combines self-supervised, supervised, and policy optimization objectives into a single sequence modeling objective, significantly enhancing FDM's generalization performance across all task types.

It is important to note that FDM employs a single multi-task network architecture, rather than a single network with shared hidden layers for all tasks. A single multi-task network implies that all tasks share the same input and output structure and use the same learning objectives, providing superior cross-domain multi-task generalization ability. In previous work, it was common to use the same policy architecture across tasks with different learning parameters for each task, leading to a loss of generality. To further improve generalization, researchers attempted to use the exact same policy on a single domain and achieved good performance on Atari57~\citep{badia2020agent57} and DMLab~\citep{espeholt2018impala}. Such methods have good generalization ability for similar tasks in the same domain but struggle with tasks from other domains. To address cross-domain multi-task problems, FDM learns a single network with the same weights across a diverse set of tasks. Similar choices have been adopted by TD and TT, demonstrating that large sequence foundation models can handle such cross-domain decision-making tasks.

\subsection{Sim-to-Real Decisions}
Many IDM models are trained in simulation environments, but real-world applications also require interaction with the physical world. Virtual environment simulations enable faster and safer training of decision models. However, once an IDM model is trained, it must be transferred to the real world, a process known as sim-to-real transfer~\citep{zhao2020sim}. When virtual environments accurately simulate real-world conditions, the transferred policy can function effectively~\citep{salvato2021crossing}. Regrettably, high-fidelity simulations demand substantial computational resources. In practice, less accurate simulators are often preferred, introducing a gap between training virtual environments and real-world applications. This necessitates IDM models being able to tolerate environmental changes from the simulator to the physical world.

Owing to their high learning capacity, foundation models like FDM can learn from various data sources, dramatically improving transferability to real-world environments. Although several attempts have been made in the literature to create sim-to-real transferable controllers, often associating robustness with sim-to-real transferability, the lack of task uniformity prevents determining which solution is most appropriate for addressing the sim-to-real gap. Previous sim-to-real methods typically require two successive training phases, which may exhibit low efficiency~\citep{salvato2021crossing}. Furthermore, if these variations and disturbances do not accurately represent the simulator's discrepancies from reality, sim-to-real transfer may fail. In FDM, given a distribution over tasks, the model learns an adaptive policy that maximizes the expected reward for a new task from the distribution. Consequently, FDM presents a promising and robust solution for quickly adapting the experience gained in real-world simulations.


\section{FDM Learns to Make Decisions Efficiently}
\label{sec:efficient}

FDM, through pre-training on offline datasets, circumvents data inefficiency concerns during learning and maintains low-latency characteristics during inference. Reinforcement Learning (RL) and Operations Research (OR) are two prominent IDM approaches for tackling distinct decision-making tasks. On the one hand, Deep Reinforcement Learning (DRL) has seen considerable success in video games~\citep{mnih2013playing}. However, DRL is plagued by sample efficiency problems due to the need for extensive, high-quality, and affordable data, which can be easily acquired from online simulators. The real world, unlike a simulator, does not provide cheap, high-quality, large-scale data. Consequently, DRL struggles to learn effective decision policies outside of simulators. On the other hand, OR algorithms are commonly employed for planning and scheduling decisions. However, OR models are inherently sensitive to problem complexity, making it challenging to implement dynamic scheduling and planning in many real-world tasks.

This section examines how FDM addresses data inefficiency issues for reinforcement learning and multi-agent learning tasks. Additionally, we describe how FDM learns OR optimization by learning OR models and transforming them from long-cycle decision services to instantaneous decision tools.

\subsection{Offline Learning from Demonstrations}
Typically, FDM is trained offline using demonstration data, adopting a highly sample-efficient imitation learning (IL) approach. Imitation learning seeks to learn from expert demonstrations to reproduce optimal behavior. IL techniques, such as behavior cloning (BC) and inverse RL (IRL), have proven effective in video games, robotic simulations, and object manipulation~\citep{zheng2021imitation,GAIL}. BC mimics expert behavior and performs well in environments with fixed parameters. IRL, which combines behavior cloning and prediction, achieves higher data efficiency and performs well in some dynamic environments. Despite these successful cases, IL still faces challenges in learning diverse behaviors, utilizing sub-optimal demonstrations, or following language/voice instructions for imitation~\citep{hua2021learning}. FDM not only inherits the benefits of IL but also naturally handles multi-modal data and multi-task learning. Moreover, it combines behavior cloning and prediction to address these challenges. FDM also bridges offline imitation and online learning, enabling consistent and efficient learning of new tasks.

\subsection{Muli-agent Decisions}
FDM employs a sequential update scheme for multi-agent decisions, effectively transforming multi-agent joint policy optimization into a sequential modeling problem. Managing non-stationary dynamics is a critical challenge in multi-agent decision learning research, rendering it difficult to apply single-agent decision models directly in multi-agent contexts. FDM overcomes this issue by converting agents' simultaneous decision-making into a sequential decision-making process for each agent~\citep{wen2022multi}. In Multi-Agent Reinforcement Learning (MARL), the outcome of actions is not independent due to agents' interactions, unlike single-agent scenarios~\citep{gronauer2022multi}. To illustrate multi-agent interactions, consider an adaptive robot car navigating an intersection. The robot car can move by steering, accelerating, or braking. Its objective is to safely traverse the intersection and reach its destination. In addition to detecting environmental factors such as positions, traffic lights, and lane markings, the robot car must be aware of other cars, including human-driven cars or other adaptive cars. With multiple adaptive agents in a shared environment, they simultaneously improve their policies, causing the environment to be non-stationary from a single agent's perspective. Sequential update schemes have been employed in Multi-Agent Decision Transformer (MADT)\citep{meng2021offline} and Multi-Agent Transformer (MAT)\citep{wen2022multi} for efficient and stable multi-agent learning. Based on the Multi-Agent Advantage Decomposition theorem~\citep{kuba2021trust}, FDM guarantees joint performance improvement and learns an optimal joint policy efficiently.

\subsection{Learning to Optimization}

FDM approaches a series of white-box OR optimization problems as sequence modeling problems, enabling highly efficient inferences once it learns a suitable policy. OR optimization algorithms find the optimal solution within certain constraints based on accurate descriptions and characterizations of specific problems~\citep{winston2004operations}. OR algorithms do not require extensive data, and their results possess strong interpretability. As such, they have been widely employed in planning, scheduling, and coordination tasks. However, the computational complexity of traditional OR algorithms increases exponentially with problem size. In real-world applications, which often involve numerous uncertainties and large problem sizes, traditional discrete optimization algorithms may struggle to provide optimal solutions within a reasonable time frame.
FDM is a data-driven approach capable of addressing the efficiency issues of OR algorithms. Once it learns a suitable approximate policy for OR problems, FDM can generate decisions with a single-pass forward inference. This means that, for any learned discrete optimization problem, FDM can eliminate computational complexity issues and achieve linear time complexity.

\section{Real-world Applications}
\label{sec:app}

In previous sections, we discussed the advantages of FDM. To further demonstrate the effectiveness of FDM in real-world IDM applications, we present three decision problems that are well-suited for FDM to address.

\subsection{Game AI}

FDM enhances the generalization capabilities of game AI due to its strong multi-modal and multi-task performance. Deep RL (DRL) methods have achieved significant advancements in game AI, with examples such as the DQN agent~\citep{DRQN}, AlphaGo~\citep{alphago}, Libratus~\citep{brown2017libratus}, OpenAI Five~\citep{berner2019dota}, and AlphaStar~\citep{alphastar}, where DRL agents have defeated professional human players. These successes indicate DRL's capacity to solve highly complex games~\citep{yin2021ai}. However, these DRL agents are typically designed for specific games.
For instance, AlphaGo Zero~\citep{silver2017mastering} employs Monte Carlo tree search, self-play, and deep learning to defeat numerous professional Go players, showcasing powerful techniques for larger perfect information games. Utilizing self-play, deep reinforcement learning, and continuous transfer via surgery, OpenAI Five became the first AI to defeat world champions in an esports game, demonstrating practical techniques for complex, imperfect information games. In contrast, the real world comprises a myriad of games, making it nearly impossible for DRL agents to discover a universal policy through self-play, deep learning, and reinforcement learning alone.
We argue that previous training frameworks, particularly self-play with distributed learning, may not be suitable for such situations, as two-player asymmetric games have distinct strategies for each side, and self-play-based mechanisms might not perform well. Through offline training, FDM can distill expert knowledge from various single-task expert agents and offer a unified solution for game AIs.

\subsection{Production Scheduling}

\begin{figure}[t!]
  \centering
  \includegraphics[width=\linewidth]{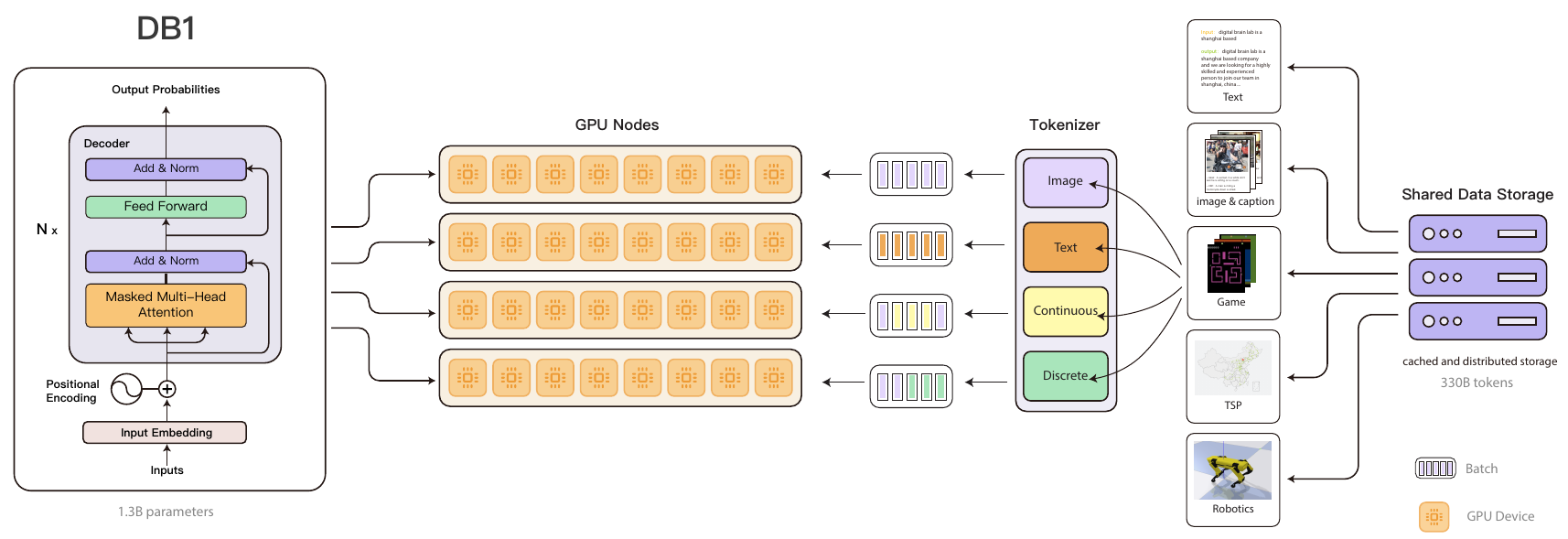}
  \caption{Training pipeline of DB1. We distribute DB1 to multiple GPU nodes and devices to perform data-parallel training. A shared data storage is equipped with caching mechanism to speed up the data loading, and a distributed strategy for data saving. The sampled training data will be discretized with 4 tokenizers before feeding to the model.}
  \label{fig:db1}
\end{figure}
FDM significantly enhances problem-solving efficiency for production scheduling, enabling the generation of feasible solutions for real world tasks instantly. Scheduling is a highly complex aspect of production system management and control, crucial for improving the rapid response and global optimization capabilities of intelligent manufacturing systems. Although numerous studies have explored various scheduling models and methods, challenges persist in the field, posing questions for future development trends.
Currently, information and intelligent technologies are driving changes and innovations in products and production organizations, increasing the complexity of manufacturing system management and control~\citep{jiang2022evolution}. As production data grows exponentially, data-driven IDM models support service scheduling in sustainable manufacturing environments that synergistically optimize production efficiency and energy consumption~\citep{waubert2022reliability}.

\subsection{Robotics}
Owing to its highly complex model, FDM can learn from diverse datasets, reducing the sim-to-real gap and enhancing the robustness of robotics in real-world applications. Achieving robots with the full range of physical capabilities that humans possess is a distant goal-arguably even more so than attaining AI systems with the full range of intellectual capabilities. In the physical world, an embodied agent encounters numerous changing factors, including physical parameters, action spaces, tasks, scene visual appearances, object geometry and topology, and more. Additionally, many critical real-world tasks demand generalizable policy learning, such as visual navigation, object manipulation, and autonomous driving~\citep{bonin2008visual}. Therefore, learning generalizable policies is essential for developing intelligent embodied agents in the real world. Most current AI robotics models lack real-world experience due to sample efficiency issues and sim-to-real gaps, which limit the capabilities of IDM models. Generalization is also crucial when robots interact with the physical world. The key challenge is collecting the appropriate data and effectively learning from it~\citep{hua2021learning}. We argue that this is the reason we require an IDM model that can adapt to new embodiments and learn new tasks with less data~\citep{singh2021reinforcement}. FDM can fulfill the generalization requirements for both embodiments and the physical world. FDM learns generalizable policies in the physical world through synergistic efforts across vision, learning, and robotics fields.

\section{Case Study: the DigitalBrain Model}
\label{sec:case}
As discussed in previous sections, FDM presents a promising solution for real world IDM applications. We trained the DigitalBrain model, denoted as DB1, an implementation of FDM that demonstrates the potential of FDM for text, image-text, video games, robotics, and operations research tasks. In the following sections, we introduce the details of DB1, including model design, dataset collection and processing, training schemes, and empirical results.

\subsection{Dataset Collection and Processing}
We introduce three types of decision tasks in our case study: \textit{simulated control}, \textit{vision \& language}, and \textit{traveling salesman problem (TSP) tasks}. For each category, we consider multiple heterogeneous sub-task sets to satisfy multi-modality learning requirements. Regarding data collection, we must ensure that the dataset is expert, as the loss function implies the training scheme of DB1 is supervised learning. To obtain high-quality datasets, the experts on simulated control tasks are implemented as well-trained reinforcement learning algorithms; the experts on vision and language tasks are humans; and the experts on TSP tasks are well-designed solvers. For data processing, we follow the same principle as Gato, i.e., quantifying and discretizing raw data into integers. We provide more details below.

\subsubsection{Simulated Control Tasks}
The simulated control tasks consist of various video games and robotics control task sets, which have been treated as common benchmarking environments for reinforcement learning. The multi-modality property in these tasks is not only reflected in the task types but also in the state/action space definition.

\begin{table}[t!]
\caption{Datasets of simulated control and TSP task suites. For each task suite, we summarize the number of tasks, the number of collected trajectories, and the number of tokens. To avoid overfitting to a large dataset, the sample weights of task suites are well-tuned instead of being proportional to the dataset size.}
\centering
\begin{tabular}{lllll}
\toprule
Task Suite            & Tasks & Trajectories & Tokens  & Sample weights\\
\midrule
DeepMind Lab      & 411         & 3M                    & 171B   & 13\% \\

ALE Atari         & 52          & 32K                        & 1.4B  & 13\% \\

Sokoban           & 1           & 910              & 2.8M                 & 1.7\% \\

BabyAI            & 55          & 5.5M           & 33.5B     & 13\% \\

DM Control Suite  & 25          & 675K            & 20.2B     & 5.2\% \\

Meta-World        & 42          & 88.2K             & 1.9B      & 11.7\% \\

Procgen Benchmark & 16          & 1.6M               & 7.2B        & 6.5\% \\

Modular RL        & 38          & 843.6K          & 13.9B     & 10.4\% \\

TSP               & 230         & 2.3M          & 108.5B    & 11.7\% \\
\bottomrule
\end{tabular}%
\label{tab:dataset}%
\end{table}%

The simulated control tasks comprise various video games and robotics control task sets, which have been treated as common benchmarking environments for reinforcement learning. The multi-modality property in these tasks is not only reflected in the task types but also in the state/action space definition.

\begin{itemize}
\item DeepMind Lab~\citep{dmlab} is a 3D learning environment developed for the research and development of  artificial general intelligence and machine learning systems. It provides a suite of challenging 3D navigation and puzzle-solving tasks for learning agents. We split the existing levels in DeepMind Lab into 6 task sets and train a v-trace agent~\citep{schmitt2020off} on each task set for data collection, 65,000 episodes per level, for a total of 1.5M episodes.

\item Procgen~\citep{cobbe2019procgen} consists of 16 procedurally generated game-like environments for evaluating sample efficiency and generalization in reinforcement learning. The data is collected with a v-trace or r2d2 agent~\citep{schmitt2020off} in each environment. All the distribution modes are set to hard, except for mazes and heists which are easy. We collect 100,000 episodes for each environment, for a total of 1.6M episodes.

\item Sokoban~\citep{racaniere2017sokoban} is a planning problem in which the agent has to push boxes to target locations. Since some of the moves are irreversible, mistakes can make the puzzle unsolvable. Planning ahead of time is therefore necessary to succeed on this puzzle.
We take advantage of the reversed play processes when generating the maps. We use a breadth-first search in the reversed play processes and extract the solutions from it as the expert data. So this expert data is optimal in most cases.

\item BabyAI~\citep{chevalier2018babyai} is a grid-world environment whose levels consist of instruction-following tasks described by a synthetic language. We generate data for these levels with the built-in BabyAI bot. The bot has access to extra information for executing optimal solutions. See Appendix C of Chevalier et al.~\citep{chevalier2018babyai} for more details about the bot. We collect 100,000 episodes for each level, for a total of 5.5M episodes.

\item Modular RL~\citep{huang2020modular_rl} is a collection of MuJoCo~\citep{todorov2012mujoco} based continuous control environments, composed of variants of the OpenAI Gym~\citep{brockman2016openai}. Each variant is a morphological modification of the original body: the set of morphologies is generated by enumerating all possible subsets of limbs and keeping only those sets that a) contain the torso and b) still form a connected graph. This results in a set of variants with different input and output sizes, as well as different dynamics from the original morphologies. We use the qpos and qval as observations, just like in the OpenAI Gym. And we train a D4PG agent~\citep{barth2018distributed} on each variant. 843.6K episodes are collected for this task suite.

\item Meta-World~\citep{yu2019metaworld} is a suite of environments for benchmarking meta-reinforcement learning and multi-task learning. We collect data from all training and test tasks in the MT1 mode by training an APPO agent~\citep{berner2019dota} from sample-factory repository~\citep{petrenko2020sample}. For this task suite, we collect 88.2K episodes in total.

\item The DeepMind Control Suite~\citep{tunyasuvunakool2020dmc} is a set of physics-based environments. For each task in the control suite, we collect two disjoint sets of data, one using only state features and the other using only pixels. We use a D4PG agent~\citep{barth2018distributed} to collect data from tasks with state features and an APPO agent~\citep{berner2019dota} to collect data using pixels. For this task suite, we collect 675K episodes in total.
\end{itemize}

\subsubsection{Vision and Language Tasks}
DB1 is trained on an extensive collection of English language resources derived from web pages. Additionally, vision-language tasks such as Microsoft COCO Caption\citep{chen2015mscoco} are included.

\subsubsection{TSP Tasks}

To investigate DB1's capabilities in handling optimization problems in operations research, we selected the Traveling Salesman Problem (TSP)~\citep{gavish1978travelling}, a well-known combinatorial optimization problem, as a representative task. Generally, TSP is modeled as a sequential problem (i.e., using coordination and neighbor information as states and node choice as decision-making), with various reinforcement learning methods developed for its solutions. Our dataset comprises 2 million trajectories from 200 tasks of different scales, specifically including 200 randomly generated tasks with 100 or 200 nodes uniformly sampled from a 2D plane. The expert for data collection in TSP is derived from LKH~\citep{lin1973effective}, a heuristic algorithm. Since the original action space is large, we further narrow DB1's action space by using the expert's closest neighbor in each decision. For state representation, we encode neighbor information using a pre-trained Graph Convolutional Network (GCN)~\citep{kipf2017semisupervised}.

Sample weights for TSP tasks and simulated control tasks are listed in Table~\ref{tab:dataset}, totaling 87\%. The remaining data consists of 8.7\% from the language corpus and 5.1\% from COCO Caption~\citep{chen2015mscoco}.

\subsection{Model Design}

As shown in Figure~\ref{fig:db1}, we train an FDM model, DigitalBrain (DB1), with approximately 1.3B parameters. DB1 uses TransformerXL~\citep{dai2019transformer} as its backbone, which is a decoder-only model containing masked multi-head attention and feed-forward networks with residual connections. Compared to GPT~\citep{brown2020gpt3,radford2019gpt2}, TransformerXL, located on the far left of Figure~\ref{fig:db1}, employs a relative position encoding scheme instead of an absolute one and introduces a memory mechanism that enhances computational efficiency during inference and theoretically increases DB1's sequence length by caching and propagating each layer's hidden state. Although DB1 only uses the memory mechanism during the evaluation phase, we find it more efficient for sampling long sequences.

As shown in Table~\ref{tab:traning+setting}, DB1 comprises 16 transformer blocks with 16 heads, 2048 layer width, and shared embedding. We use a dropout probability of $0.1$, PreNorm~\citep{xiong2020layer} for layer normalization, and GeGLU~\citep{shazeer2020glu} as the activation function.

\begin{table}[t!]
\centering
\caption{Hyperparameters of DB1.}
\begin{tabular}{ll}
\toprule
Hyperparameter      & Value/Choice \\
\midrule
Transformer blocks  & 24          \\
Attention heads     & 16          \\
Layer width         & 2048        \\
Shared embedding    & True        \\
Layer Normalization         & PostNorm        \\
FeedForward size before activation & 8192 \\
Activation Function & GeGLU       \\
Dropout Probability & 0.1      \\
\bottomrule
\end{tabular}%
\label{tab:traning+setting}%
\end{table}%
\subsubsection{Loss Function} 
We use a classical language modeling loss in the setting of teacher forcing~\citep{williams1989learning}, i.e., we only predict the next token during training. For a batch of $N$ instances of sequence length $L$, the loss is computed as the following equation with a given model parameter $\theta$:
\begin{equation}
    \mathcal{L}(\theta) = - \sum\limits_{b=0}^{N-1}\sum\limits_{l=0}^{L-1}m(b, l)\log p_\theta(s_l^{(b)}| s_0^{(b)}, s_1^{(b)}, \ldots, s_{l-1}^{(b)}),
\end{equation}
where $s_l^{(b)}$ denotes the $l$-th token in the $b$-th sequence of a batch, $m(b,l)$ denotes a binary loss mask on each token. We set $m(b, l)=1$ when $s_l^{(b)}$ is a token of an action or text, which means we compute the loss for all text tokens and action tokens in RL tasks, equivalent to Gato.

\subsubsection{Tokenization for Multi-modal Data} 
Given the introduction of numerous multi-modal and heterogeneous datasets and tasks during training, it is necessary to convert these different data types into a unified representation. We implemented four types of tokenizers for different data types, as follows:

\begin{itemize}
  \item Text. A tokenizer is first trained on the collected corpora with a fixed vocabulary size of 32k. For the OpenWebText dataset, we tokenize it beforehand and build a sample index based on sequence length during training.
  \item Image. Images are initially divided into a series of 16-by-16 patches, followed by normalization within each patch. Without tokenization, each patch is mapped into the embedding space of our model. We use a ResNet-v2 block, similar to Gato, for extracting visual attributes. GroupNorm with 32 groups is used instead of BatchNorm. Additionally, the patch positional encoding specified in Gato is employed to inform the model about a patch's global position within the picture it was extracted from. By normalizing the patch's pixel intervals by the picture resolution, the relative row and column intervals of the patch are determined. The row and column normalized intervals are quantized into a vocabulary size of 128, used to index a row and column table of learnable position encodings. The transformation of quantized row and column intervals into indices depends on whether the model is being trained or evaluated: during training, a random index is uniformly selected from the quantized interval, while during evaluation, we deterministically take the (rounded) mean of the interval. After retrieving row and column position encoding from the embedding table, it is added to the vision embedding created by a ResNet-v2 block.
  \item Discrete data: Discrete values are tokenized into the range [0,1024), overlapping with our text vocabulary.
  \item Continuous data: Continuous values are tokenized into the range [32000, 33024). A special token $\varphi$ is introduced as a split between each $s_t$ and $a_t$, with token ID 33204, represented as $(s_t, \varphi, a_t)$.
\end{itemize}

For any given environment, its types are categorized into text, image, and discrete and continuous values. For nested structures, each underlying attribute is categorized. Then, each observation within a timestep is concatenated according to a fixed order. First, values of different types are ordered as text, vision, and tensors. Vision patch tokens follow raster order, while tensor tokens are in row-major order. For nested structured input with each input type, the inputs are arranged lexicographically by their keys. To distinguish between the observation and action parts, we additionally use a specific reinforcement learning local positional encoding, as described in Gato. 

Upon completing tokenization and sequencing, a parameterized embedding function is applied to each token, encompassing both observations and actions, in order to generate the final model input. 
For tokens associated with text, discrete- or continuous-valued observations, or actions for any given time-step are embedded through a lookup table into a learned vector embedding space. Learnable position encodings are incorporated for all tokens based on their local token position within the respective time-step. Tokens corresponding to image patches for any time-step are embedded utilizing a single ResNet block, which yields a vector for each patch. In the case of image patch token embeddings, a learnable within-image position encoding vector is also incorporated.

\subsubsection{Comparisons with Gato}
DB1 primarily focuses on the replication and validation of Gato, and attempts to make improvements in two aspects: network structure and parameter size, as well as task types and the number of tasks:

Parameter size and network structure: DB1 has $1.3$ billion parameters, striving to be as close as possible to Gato in terms of parameter size. Overall, the structure employed by the DB1 is similar to that of Gato (same number of Decoder Blocks, hidden layer size, etc.). However, in the FeedForwardNetwork, due to the GeGLU activation function introducing an additional $1/3$ of the parameter size, the DB1 uses a $4 * n\_embed dimensional$ hidden layer state that, after passing through the GeGLU activation function, becomes a $2 * n\_embed dimensional$ feature to approach Gato's parameter size. In other aspects, we share embedding parameters at the input and output encoding ends, as in Gato's implementation. Unlike Gato, we opted for the PostNorm approach in the selection of layer normalization. We also employed mixed-precision computation in Attention, which improves numerical stability.

Task types and the number of tasks: The number of experimental tasks in DB1 reaches 870, an increase of $44.04\%$ compared to Gato, and a $2.23\%$ improvement in expert performance ($>=50\%$). In terms of specific task types, DB1 mainly inherits Gato's decision-making, image, and text-based tasks, with the number of tasks in each category remaining essentially consistent. However, in the area of decision-making tasks, DB1 introduces more than $200$ additional real-world scenarios, namely, solving the Traveling Salesman Problem (TSP) with $100$ and $200$ node scales. These tasks involve randomly selecting $100-200$ geographical locations from major cities in China as node representations.

\subsection{Training Scheme}
\textbf{Hyperparameters.} As shown in Table~\ref{tab:traning+setting}, DB1 consists of 16 transformer blocks with 16 heads, 2048 layer width and shared embedding.
We use dropout probability of $0.1$, PreNorm~\citep{xiong2020layer} for layer normalization and GeGLU~\citep{shazeer2020glu} as activation function.  

\begin{figure}[t!]
    \centering
    \includegraphics[width=1.\linewidth]{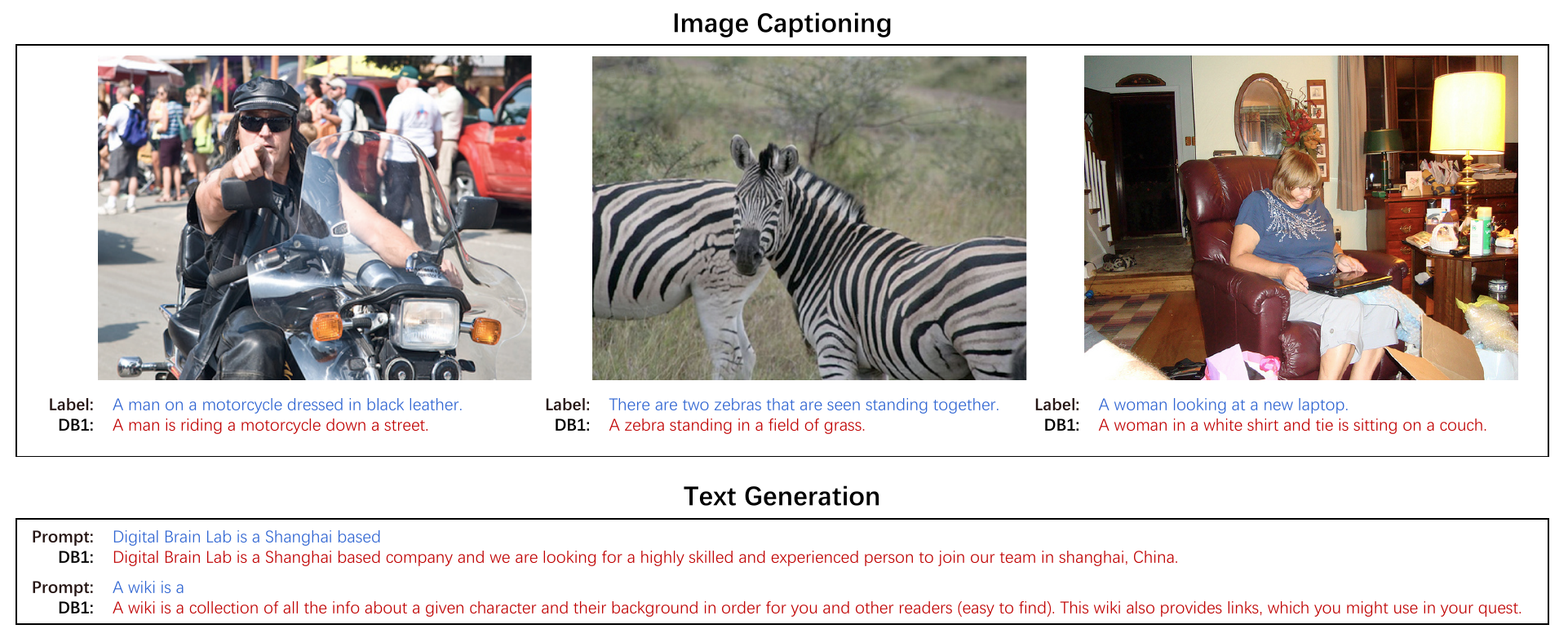}
    \caption{Selected results of vision and language tasks.}
    \label{fig:eval text and image}
\end{figure}

\begin{figure}[t!]
    \centering
    \includegraphics[width=0.8\linewidth]{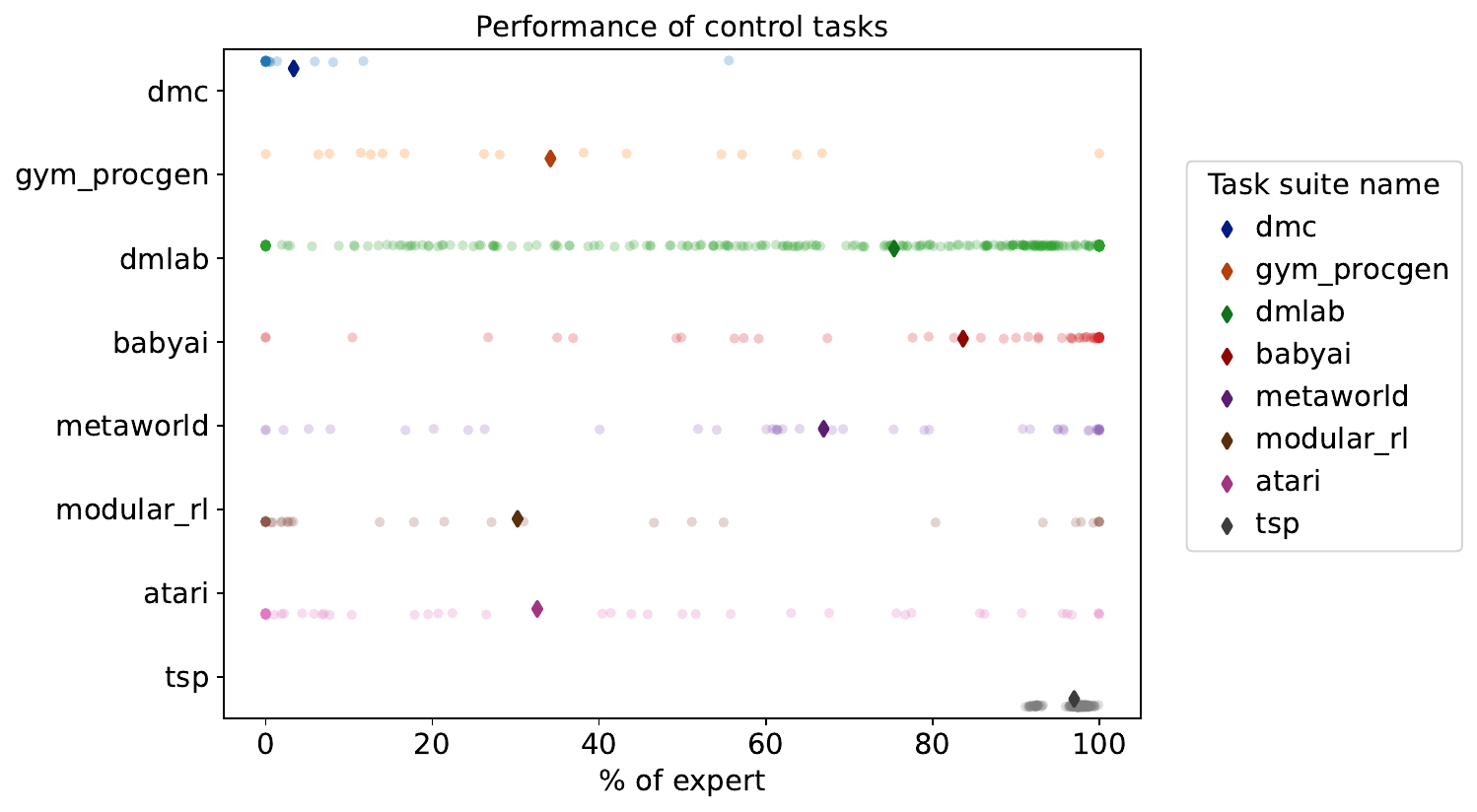}
    \caption{\textbf{Performance of DB1 on simulated control tasks.} The performance of a single task is plotted in each row organized by domain. Numbers on the $x-$axis reflect a certain percentage of expert score, with $0$ denoting the performance of a random agent. Each dot on the figure represents a certain task. The diamond means the mean score of each domain.}
    \label{fig:eval_overall2}
\end{figure}

\textbf{Optimizer.} The AdamW optimizer~\citep{loshchilov2017adamw} is used, with linear warm-up and cosine schedule decay. The linear warmup lasts for $10,000$ steps, starting with a zero learning rate and ending at a maximum learning rate of $1e^{-4}$. This learning rate is then cosine decayed by a factor of $10\times$ over $1,000,000$ steps. The AdamW optimizer has parameters $\beta_1=0.9, \beta_2=0.95$, $\epsilon=1e^{-8}$. We use a batch size of 512 and a sequence length of 1024 tokens for all models.

\textbf{Prompt Learning in Control Tasks.} DB1 is trained to complete multiple control tasks by modeling them as different sequences. However, for tasks sharing the same observation and action specifications, more information is needed for the model to distinguish them. Compared to one-hot identifiers, DB1 is trained with prompt conditions~\citep{liu2021ppprompt}.


\textbf{Dataset Building and Cache.}
%
The preprocessing process for reinforcement data is similar to that used in the Trajectory Transformer~\citep{trajectory-transformer}. We operate on separate trajectories split by environmental terminal signals; then, episodic returns, path lengths, and other statistical properties are calculated. Our index is built by enumerating a 3-tuple of path index, start index, and end index on a sliding window that contains the minimal timestep, from which we can obtain tokens of at least the sequence length set for the model.

Loading all data into memory at once is usually infeasible due to its volume. To load indices and processed data as needed, we save them on disk for each dataset.

During training, we sample a batch of 512 sequences from different task domains with a fixed sample weight defined before training. We employ a prompt scheme similar to the one mentioned in Gato~\citep{reed2022generalist}: 25\% of samples in a batch will be prepended with a prompt sequence. Half of these prompt sequences come from the end of an episode for goal conditioning, and the other half are subsequences uniformly sampled from an episode. We use a fixed length of prompt length, set to about half the length of the model's input sequence. During evaluation, we select a successful demonstration as a prompt, i.e., episodes with top 10\% returns, and feed the first $k$ timesteps, whose length after tokenization is just equal to or more than the model's input sequence.

\subsection{Evaluation Results}

We assessed DB1's performance after 250,000 training iterations. For each simulated control and TSP task, we normalized the demonstration results to a range between 0 and 1. Specifically, we normalized the results using the random policy score $R_{\min}$, expert score $R_E$, and the formula: $score= \frac{1}{N}\sum^N_{i=1}(r_i - R_{\min})/(R_E - R_{\min})$. To account for the influence of randomness, we conducted $N=10$ evaluations for each task ($r_i$ in the equation represents the result of the $i$-th evaluation) and reported the average normalized score as DB1's performance.

Figure~\ref{fig:eval_overall2} presents the performance distribution for each task set in the control and TSP tasks. Statistically, DB1 surpassed 50\% of the expert scores in 661 of the 870 tasks. In BabyAI~\citep{chevalier2018babyai}, DB1 achieved approximately 90\% of the expert score across all 55 tasks. DB1 attained roughly 96\% of the expert score for all TSP tasks, which may indicate that for combinatorial optimization problems like TSP, achieving a reasonably good result is relatively easier than obtaining a near-optimal one. The performance in DM Control Suite~\citep{tunyasuvunakool2020dmc} and Atari~\citep{ale-atari} was not as strong, potentially because the collected data were insufficient for supervised learning alone.

For the image-caption evaluation, we randomly sampled images from the COCO Caption Dataset~\citep{chen2015mscoco} and presented the results without cherry-picking. Figure~\ref{fig:eval text and image} displays selected results for text-generation and image-captioning tasks. The results demonstrate that DB1 effectively learned from the training corpus and established connections between vision and language modalities. The caption sentences indicate that DB1 can identify the primary element in an image. Although minor errors exist in details such as numbers and colors, a larger dataset could potentially address these issues.

We have not directly compared the performance of DB1 with the GATO model, as the GATO model, its training tasks, and the collected training data are proprietary, preventing us from reproducing the results. In DMC and Atari tasks, the performance of DB1 is inferior to that of GATO, as the expert models we utilized for data collection were not sufficiently strong and could not achieve the effects of the algorithms employed by DeepMind (which are not officially open-source). Since the primary training method for DB1 and GATO is behavior cloning, the quality of the data essentially determines the upper limit during actual testing.

\subsection{Generalization Capabilities}

We believe that the generalization capabilities of DB1 indeed vary across different tasks. This is because, firstly, the model itself possesses different generalization abilities for different tasks, and secondly, the inherent characteristics of different tasks determine whether they are easily generalizable or not. For instance, in BabyAI, which is an environment built on a minigrid, the dynamics of the environment are essentially fixed. However, the requirements for different tasks are represented by natural language. As both DB1 and GATO have been pretrained on natural language, they are inherently well-suited to generalize in such settings. In contrast, Atari tasks are more difficult to generalize, and this issue has been alleviated only slightly after incorporating prompt conditioning. Due to the insufficient training length of the models and the longer horizon of the environment itself, generalization remains challenging for unseen tasks.

In addition, online adaptation for novel tasks can be employed by utilizing offline pre-trained models. The most elementary optimization technique entails fine-tuning or specialization for both supervised and reinforcement learning paradigms. Numerous implementations exist for this objective, such as refining the model concurrently with learning to predict actions. Consequently, we can accomplish model-based planning and self-bootstrapping style training.
Moreover, when reward labels are available, it is possible to instruct the model to assimilate values akin to return-to-go. This approach aids the model in actively discerning the quality of the training data and sampling in accordance with the specified target return. In scenarios with significant environmental randomness, techniques such as adversarial training and contrastive learning may be deployed to mitigate the influence of environmental stochasticity on the encoding of future information.

\section{Conclusion}
\label{sec:con}
FDM is a fundamental yet efficient model for various multi-modal, multi-task decision-making problems, encompassing text, vision, reinforcement learning, and operations research tasks. As a foundation model, FDM inherently possesses strong generalization and efficient learning capabilities. In accordance with the scaling law, FDM's performance will improve as the scale of parameters and data increases, further enhancing the cross-task learning and generalization abilities of the model. Future work will involve developing a more autonomous decision-making model and integrating state-of-the-art IDM models into real-world environments by continuously training, fine-tuning, and scaling up the FDM.




\newpage








\vskip 0.2in
\bibliography{Reference}

\end{document}